\documentclass{article}
\usepackage{spconf,amsmath,graphicx}
\usepackage{url}
\newcommand\norm[1]{\left\lVert#1\right\rVert}

\title{3D Point Cloud Registration with Shape Constraint}
\name{Swapna Agarwal, Brojeshwar Bhowmick}
\address{Embedded System and Robotics, TCS Research \& Innovation, India}
\begin{document}
%
\maketitle
\begin{abstract}
In this paper, a shape-constrained iterative algorithm is proposed to register a rigid template point-cloud to a given 
reference point-cloud. The algorithm embeds a shape-based 
similarity constraint into the principle of gravitation. The shape-constrained gravitation, as induced 
by the reference, controls the 
movement of the template such that at each iteration, the template better aligns with the reference 
in terms of shape. 
This constraint enables the alignment in difficult conditions indtroduced by change (presence of outliers and/or missing parts), translation, rotation and scaling. We discuss efficient implementation techniques
with least manual intervention. The registration is shown to be useful 
for change detection in the 3D point-cloud.
The algorithm is 
compared with three state-of-the-art registration approaches. The experiments are done on both 
synthetic and real-world data. 
The proposed algorithm is shown to perform better in the presence of big rotation, structured and unstructured
outliers and missing data.
\end{abstract}
 \vspace{-0.2cm}
\section{Introduction}
\label{I}
\begin{figure}[!t]
  \centering
    \centerline{\includegraphics{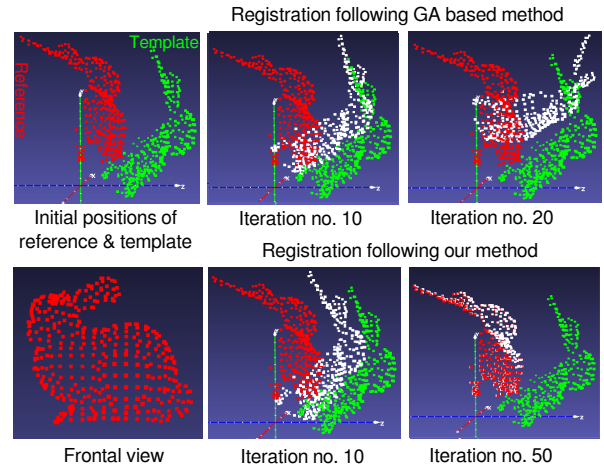}}
 \vspace{-0.2cm}
\caption{Example comparing the registration results of the proposed method vs. that of GA \cite{golyanik2016gravitational}. 
The frontal view of the bunny data shows that the lower body part is heavier compared to the head. In GA, the heavier parts of
the template (green) get attracted more by the reference (red), irrespective of shape, resulting in misalignment (first row). In our method, 
the parts of the 
template get attracted by the parts of the reference having similar shape. This gives better alignment (white) even in presence of large
rotation. In the example, the template is given a $50^\circ$ of rotation wrt the x-axis.}
 \vspace{-0.2cm}
\label{f1}
\end{figure}

 \vspace{-0.2cm}
With the increasing availability of 3D data acquisition devices such as Kinect \cite{zhang2012microsoft}, \cite{kinect2016}, 
the researchers are inclining towards exploiting the 3D information for a variety of 
processes. Registration of 3D point-clouds is a necessary initial step for many such
processes. Examples include change detection in a scene, industrial quality 
control, pose tracking \cite{ge2015sequential} etc. Generally, two point-clouds (called
reference and template respectively) of an object or scene are captured at two time instances, possibly from different camera
positions.
The scene may have gone under some changes during this time. 
The registration refers to 
the process of translating, rotating and scaling the template such that it optimally aligns with the 
reference. In this paper, we focus on
the automatic rigid registration of 3D point-clouds. The proposed algorithm does not require any color, texture, 
point correspondence or point topology
information of the point-clouds. Moreover, presence of noise, outlilers and missing parts
make the problem even more challenging. Now, we briefly review the related literatue and state our 
contributions (Fig. \ref{f1}).

One of the earliest and most popular method for rigid point set registration is Iterative Closest
Point (ICP) \cite{besl1992method}, 
\cite{zhang1994iterative}. Using 
non-linear optimization algorithm, ICP minimizes the mean squared distances between the two point sets. 
Though this method is simple to implement, it is sensitive to outliers and the performance depends 
on the initial alignment \cite{golyanik2016gravitational}. Some variants \cite{rusinkiewicz2001efficient}, 
\cite{fitzgibbon2003robust} of ICP are also there.
\begin{figure}[!t]
  \centering
    \centerline{\includegraphics{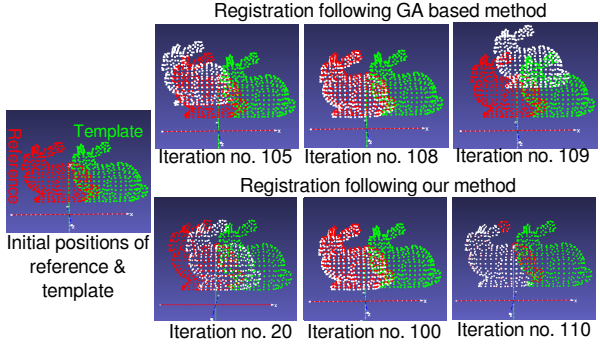}}
 \vspace{-0.2cm}
\caption{Example comparing iterative alignment results of the proposed registration method with that 
of GA \cite{golyanik2016gravitational}. 
The white point cloud shows the registraion result. The proposed algorithm reaches local minima
very fast (20 vs. 105 iterations in the second column). Near local minima, our approach takes smaller steps while
GA overshoots (third to fourth columns).}
 \vspace{-0.2cm}
\label{f2}
\end{figure}

While ICP \cite{besl1992method} assigns discrete correspondences between the points of the two sets, later 
methods such as Robust Point Matching (RPM) \cite{gold1998new} and  \cite{luo2003unified}, \cite{chui2000new}, 
\cite{chui2003new} assign soft correspondences.
 A set of methods \cite{mcneill2006probabilistic}, \cite{sofka2007simultaneous} treat the 
registration problem as maximum likelyhood estimation problem where the template points are assumed 
to be the centroids 
of Gaussian Mixture Model (GMM) and the reference points are seen as data points.  
Expectation-Maximization (EM) algorithm 
is generally used for optimization of likelyhood function. A closed form solution to the M-step of the EM algorithm 
for multidimensional case is proposed in \cite{myronenko2010point}. Their method is called Coherent Point Drift (CPD). 
The GMM based methods perform better in the presence of noise. CPD was refined in \cite{wang2011refined} 
and \cite{golyanik2016extended} for structured outliers. 
Other recent mentionable approaches are \cite{biswas2006invariant}, \cite{marques2013guided} and \cite{peng2014street}. 
Recently in 2016, a new 
Gravitational Approach (GA) 
based registration algorithm is proposed \cite{golyanik2016gravitational}. 
Under GA algorithm, the template moves towards the reference under the influence 
of gravitaional force as induced by the reference. GA based method is shown to 
perform better compared to CPD under more than 50\% uniform noise. The performance of GA decreases with 
increasing rotation (more than $45^o$) and change 
(extra or missing parts) \cite{golyanik2016gravitational}. We propose substantial modifications to the GA algorithm 
\cite{golyanik2016gravitational} to overcome the disadvantages of GA.

Our contributions are as follows. (1) GA method minimizes the 
distance between the center-of-masses of the two point-clouds but does not ensure proper shape 
alignment (see Fig. \ref{f1}). We constrain the force of
attraction with a measurement of local shape-similary for better alignment of the shapes (i.e., handles rotation and outliers
better).
(2) In GA, the force of attraction is inversely proportional to the distance beetween the 
two points involved. As a result, the increased speed at local minima (where the distance is small) 
does not allow the algorithm to
converge (see Fig. \ref{f2}). To handle this situation, a number of free parameters are introduced in GA. In our algorithm,
the force of attraction is proportional to a monotonically increasing function of distance. 
This modification allows for better
convergence without the need for extra free parameters.
(3) GA has limited capability to handle scale change \cite{golyanik2016gravitational}. 
In our method, to handle scale change, an orientation and translation invariant model is proposed that uses
the spatial point distribution of the two point-clouds. 
(4) The new algorithm is evaluated extensively for 
fully automatic performance as opposed to a number of (seven) free, manually adjusted parameters of 
GA method \cite{golyanik2016gravitational}. 

The proposed algorithm is compared with ICP \cite{besl1992method}, CPD \cite{myronenko2010point} 
and GA \cite{golyanik2016gravitational} in Section \ref{V} and the conclusions are drawn in Section \ref{VI}.
Next, we briefly present the GA based registration method following which we elaborate the proposed 
shape-constrained registration algorithm.
 \vspace{-0.2cm}
\section{Gravitational Approach}
\label{II}
 \vspace{-0.2cm}
In Gravitational Approach (GA) based registration, each point in one point set (called reference)
attracts each point in the other point set (called template). The force of attraction is governed
by the following formula.
\begin{equation}
\mathbf{f}^{Yi}=-Gm^{Yi}\sum_{j=1}^N{\frac{m^{Xj}}{(\norm{r^{Yi}-r^{Xj}}^2+\epsilon^2)^{3/2}}}\mathbf{n}_{ij} -\eta v^{Yi}
\label{e1}
 \vspace{-0.2cm}
\end{equation}
In (\ref{e1}) G is the gravitational constant and $\mathbf{f}^{Yi}$ represents the total force applied on a 
point $Yi$ of the template. The symbols $Xj$, $m^{Xj}$, $r^{Xj}$ and $N$
rerpesent the $j$th point in the reference, mass of $Xj$, absolute
coordinates of $Xj$ and number of points in the reference respectively. The symbols $\mathbf{n}_{ij}$, $v^{Yi}$ and $\eta$ represent a 
unit vector in the direction of force, the velocity of $Yi$ in the previous iteration and a constant 
respectively. The template, as a rigid body, gets displaced in an iterative way under the influence of the cumulative 
force induced by all the points in the reference. The term $\epsilon^2$ 
does not allow the force to increase beyond a certain small distance 
between the reference and the template 
(note that $\mathbf{f}^{Yi} \propto 1/(\norm{r^{Yi}-r^{Xj}}^2)$). The term $-\eta v^{Yi}$ in 
(\ref{e1}) acts as friction that controls the velocity near local minima. 
Given the force, the displacement of the template is estimated ollowing Newton's second law of motion. 
The scale and rotation, required for registration, are estimated 
using the new positions of the template points which are estimated following (\ref{e1}).
 \vspace{-0.2cm}
\section{THE PROPOSED REGISTRATION APPROACH}
\label{III}
 \vspace{-0.2cm}
The objective of the gravitational force is to minimize the distance between the center of mass of the two 
objects. On the other hand, the objective of registration is to align the two objects (point-clouds here) 
such that the parts of the objects having similar shapes map to each other. 
Therefore, we need to modifiy the principles of gravitation to accommodate this constraint. We explain these modifications next.
 \vspace{-0.2cm}
\subsection{Modified Gravitational Approach}
In GA, $\mathbf{f}^{Yi} \propto m^{Xj} m^{Yi}$. As a result, 
in registration following (\ref{e1}), the heavier parts of the point-clouds attract each other more irrespective
of the shape, resulting
in mis-alignment (e.g., Fig. \ref{f1}). Therefore, we modify (\ref{e1}) such that,
\begin{equation}
\mathbf{f}^{Yi} \propto g(s^{Xj,Yi}),
\label{e2}
 \vspace{-0.1cm}
\end{equation}
where $s^{Xj,Yi}$ represents a measure of shape similarity of local neighborboods centered at $Xj$ and $Yi$ respectively. 
$g$ is a monotonically increasing function of $s$. We have 
evaluated different features (e.g., histogram of normals, coefficients of polynomial approximating the local
surface, curvature) that can represent shape in spatially local domain. The curvature \cite{curvature2017} seems to be the best
representative of local shape. As a similarity measure of shape, any radial basis function (RBF) can be used 
where the output
decreases monotonically as the dissimilarity increases. In our implementation, we used
 \vspace{-0.1cm}
\begin{equation}
s^{Xj,Yi}=exp(-\frac{|a^{Xj}-a^{Yi}|^2}{\sigma ^2}),
\label{e3}
 \vspace{-0.1cm}
\end{equation}
where, $|.|$ represents absolute value, $\sigma$ and $a^{Xj}$ represent the spread of the RBF funtion and the curvatue value
of the neighborhood centered at $Xj$. 
In the proposed registration approach, we want the template to make large movement if the distance between the 
template and the reference is large. We want it to take tiny steps for finetuning when close to alignment. This is ensured
by the following formula.
\begin{equation}
 \mathbf{f}^{Yi} \propto h(\norm{r^{Yi}-r^{Xj}}).
 \label{e4}
  \vspace{-0.1cm}
\end{equation}
The value of the function $g$ in (\ref{e2}) or the function $h$ in (\ref{e4}) can be the value of the 
argument itself or some suitable monotonically increasing function as we shall discuss in Section \ref{IV}. Therefore, we 
propose the following formula to estimate the total force applied on 
a point $Yi$ of the template.
\begin{equation}
 \mathbf{f}^{Yi}=-G\sum_{j=1}^N g(s^{Xj,Yi})h(\norm{r^{Yi}-r^{Xj}})\mathbf{n}_{ij}.
 \label{e5}
  \vspace{-0.1cm}
\end{equation}
Note that in (\ref{e5}) we could drop two free parameters $\epsilon$ and $\eta$ as used in (\ref{e1}) in 
GA based registration \cite{golyanik2016gravitational}. 
 \vspace{-0.2cm}
\subsection{Estimating Translation, Rotation and Scale}
We estimate the displacement represented by say, $d$ of the template following Newton's second law of motion.
\begin{equation}
d=(\frac{\mathbf{f}}{m^{X}}time +v^{X})\times time.
\label{e5.5}
 \vspace{-0.1 cm}
\end{equation} 
In (\ref{e5.5}) $\mathbf{f}$, $m^{X}$ and $v^{X}$ represent the total force ($\sum_i^M\mathbf{f}^{Yi}$) applied on the 
template $X$, mass of $X$ and velocity of $X$ in a previous iteration repsectively. 

The rotation matrix $R$ is estimates following Kabsch algorithm \cite{kabsch1976solution}. Let $Y$ and $Y_D$ represent the 
mean subtracted $M\times 3$ matrices representing the $M$ coordinates of the template before and
after translation following (\ref{e5.5}). Let $\hat{C}=Y_D^TY$ represent the corss-covariance matrix. The
rotation matrix $R$ is estimated using Singular Value Decomposition (SVD) of $\hat{C}$. After SVD 
$\hat{C}=\hat{U}\hat{S}\hat{V}^T$ and let $d=sign(det(\hat{V}\hat{U}^T))$. Then,
 \vspace{-0.2cm}
\begin{eqnarray}
 R=\hat{V}
 \begin{bmatrix}
  1 & 0 & 0\\
  0 & 1 & 0\\
  0 & 0 & d
 \end{bmatrix}
\hat{U}^T .
\label{e6}
 \vspace{-0.1cm}
 \end{eqnarray}

In GA, the scale is estimated as the ratio of the positions estiamted using (\ref{e1}) and previous positions
of the template cloud points. Any error occuring in estimating the translation using (\ref{e1}) gets propagated
to estimating the scale at each iteration of GA based registration. Our model 
does not depend on the estimation of translation. We find the 
eigen valules of the covariance matrix (say $C$) of each of the two point-clouds. The largest eigen-value represents a 
measure of the lengh of the point-cloud distribution in the direction of the largest variance.This
measurement is independent of the orientation or relative position of the two point-clouds. 
Using SVD we have $ C=USV^T$ where the diagonal elements of $S$ represent the eigen values and $U$ and $V$ are orthogonal 
matrices. We estimate the scale as $c=\frac{eX}{eY}$ 
where $eX$ and $eY$ are the largest eigen values of the reference and the template respectively.
(see Supplementary Fig. S1). We scale the template ($Y$) as.
 \vspace{-0.1cm}
\begin{equation}
Y'=V diag(c) V^T Y^T(t).
 \label{e8}
  \vspace{-0.1cm}
\end{equation}
In (\ref{e8}), $t$ stands for iteration number and $diag(c)$ represents a square matrix with $c$ in the diagonal and zeros 
elsewhere. In each iteration of our algorithm, the template is updated using
 \vspace{-0.2cm}
\begin{equation}
 Y(t+1)=Y'R+\mu+d.
 \label{e9}
  \vspace{-0.1cm}
\end{equation}
In (\ref{e9}) $\mu$ is a $M\times 3$ matrix where each row represents the mean position of the template.
 \vspace{-0.2cm}
\section{Implementation Details and Discussion}
\label{IV}
 \vspace{-0.2cm}
The value of shape similarity $g(s^{Xj,Yi})$ is the main factor controlling the translation and rotation of the template. If we 
give equal weightage to all the local shapes then we set $g(s^{Xj,Yi})=s^{Xj,Yi}$. In indoor and some ourdoor scenes, non-planar
local shapes play vital role in registration compared to planar shape. Therefore, we can set $g$
as follows to give more weightage to non-planar local shapes.
 \vspace{-0.2cm}
\begin{equation}
 g(s^{Xj,Yi})=a^{Xj}a^{Yi}s^{Xj,Yi}.
 \label{e10}
  \vspace{-0.1cm}
\end{equation}
For finding the curvature ($a$), setting the radius 
of the local neighborhood to $\sqrt{eX}$ and $\sqrt{eY}$ respectively yields good results. We design $h$ of (\ref{e4}) as 
follows.
\begin{equation}
h(\norm{r^{Yi}-r^{Xj}}) = h(\norm{r^{Yi}-r^{Xj}}^p), \text{where } p \ge 1.
 \label{e11}
  \vspace{-0.1cm}
\end{equation}
Higher value of $p$ results in greater speed but oscillation about the local minimal. Lower value of $p$ results in better 
registration but makes the process slow.  
The mass $m$ of every point is 
assumed to be 1 and $v^{X}$ is set to 0. $m$ and $v^{X}$ can be modified depending on prior information, if any.
We also suggest that $G$ of (\ref{e5}) should take larger value near convergence when the 
algorithm takes smaller steps. We modify $G$ as follows.
\begin{equation}
 G(t+1)=0.1\times G1+0.9\times G1/(1+exp[-\{t-E/2\}])
 \label{e12}
  \vspace{-0.1cm}
\end{equation}
where, we set $G1=10000$ and $E=400$ is the number of iterations. For noisy real scenarios, pyramidal approach
can be taken where the size of the local neighborhood and $\sigma$ of (\ref{e3}) decreases with iteration 
(Supplementary Fig. S2).
 \vspace{-0.2cm}
\section{Evaluation}
\label{V}
\begin{figure}[!t]
  \centering
    \centerline{\includegraphics[width=3.39 in]{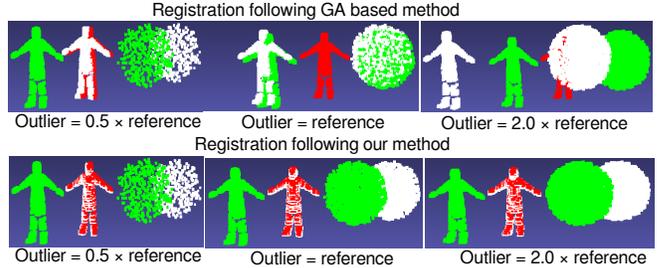}}  
 \vspace{-0.2cm}
\caption{Comparison of our method with GA \cite{golyanik2016gravitational} in the presence of structured outlier. With increasing mass of the outlier, the 
registration performance following GA degrades whereas our method consistently performs well. Red: reference, green:
template, white: template after registraion.}
\label{f4}
\end{figure}

 \vspace{-0.2cm}
\textbf{On Synthesic data:}
We evaluate the performance of the proposed method on data with different amounts and types of noise (structured and 
unstructured outliers, missing data). 
In Fig. \ref{f4} we compare our method with GA \cite{golyanik2016gravitational} in 
the presence of structured outliers. When the mass (no. of points) in the outlier is less than the 
reference (human scan), both the methods work well (Fig. \ref{f4}, col 1). As the mass of the outlier increases, 
following GA, the outlier rather than 
the human shape moves towards the reference (Fig. \ref{f4}, col 2, 3). Following our method, regardless of the mass of 
the outlier, the human shape (downloaded from \cite{humanscan} \cite{papadakis2014canonically})
in the template registers with the human shape in the reference. Here come into light the advantage of our method over GA 
\cite{golyanik2016gravitational}. Unlike GA where the movement of the template 
depends on the mass, in our method the movment depends on the shape.

\begin{figure}[!t]
  \centering
    \includegraphics[width=0.23\linewidth]{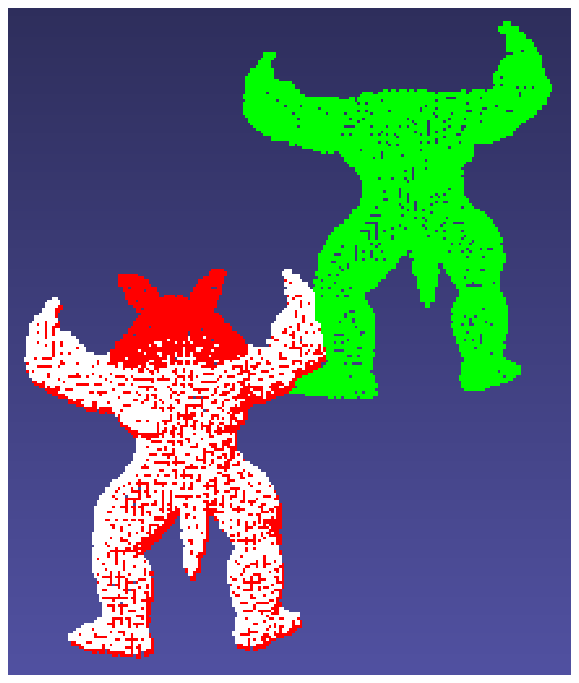}
     \includegraphics[width=0.23\linewidth]{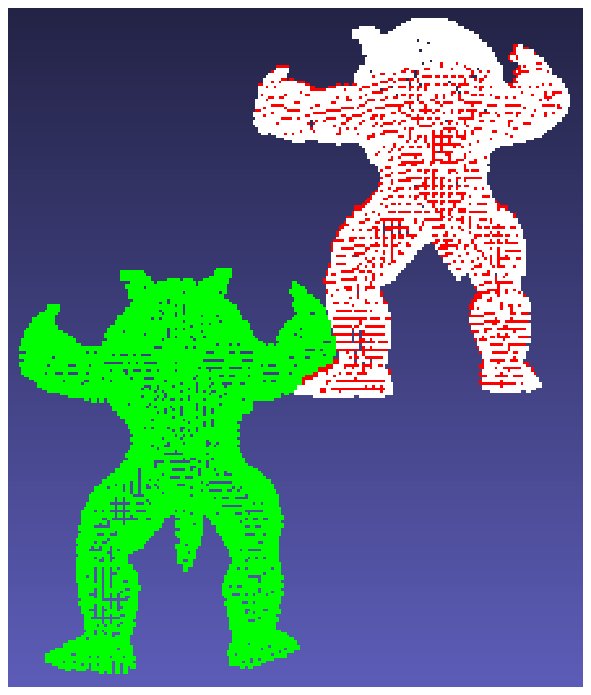}
   \includegraphics[width=0.25\linewidth]{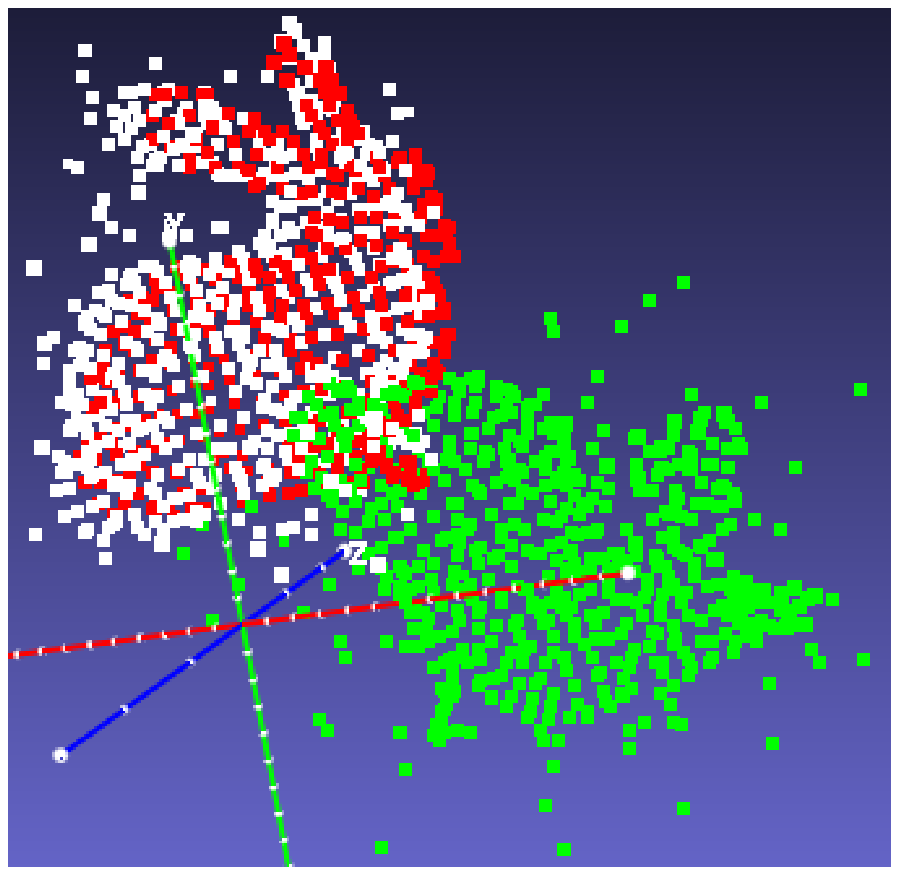}
    \includegraphics[width=0.25\linewidth]{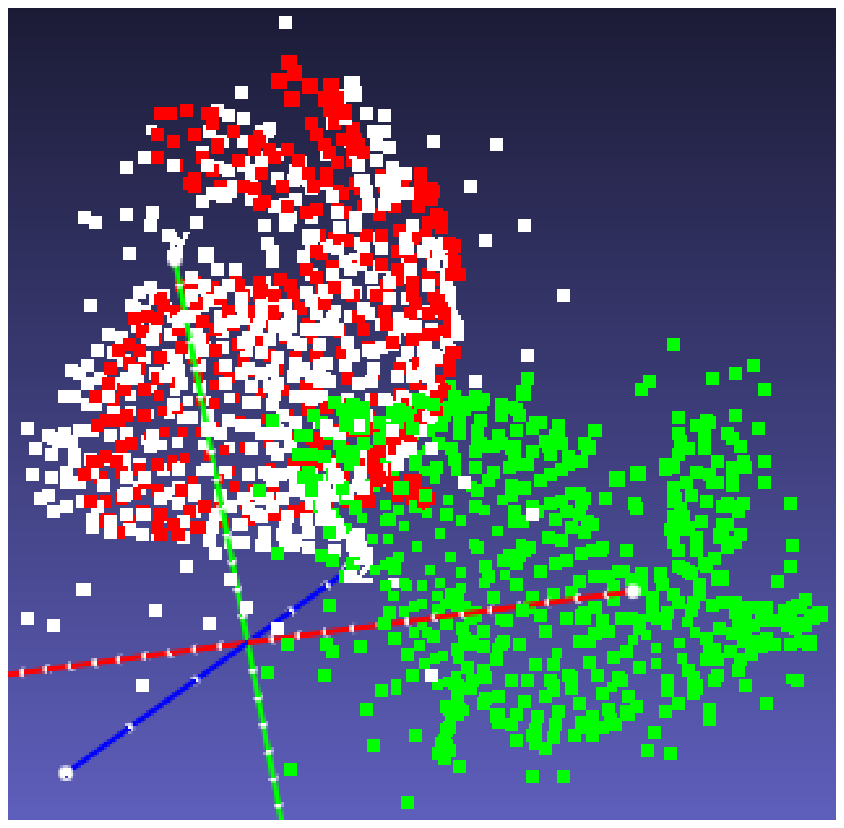}
      \vspace{-0.1cm}
 \centerline{(a) \hspace{0.23\linewidth}(b) \hspace{0.23\linewidth}(c) \hspace{0.25\linewidth}(d)}\medskip
 \vspace{-0.4cm}
\caption{Registration results of our method on point-clouds with 18\% deleted points (a, b) and 50\% outliers (c, d). 
In (a), (b), (c) and (d), the partial cloud, original full cloud, cloud with Gaussian outlier and Uniform outlier respectively
are used as the template.}
\label{f5}
\end{figure}

Fig. \ref{f5} shows the performance of the proposed method on data with missing points. We delete 18\% consecutive points 
from the head of the Armadillo \cite{stanfordbunny} data. For both
the cases when the full Armadillo or the one with missing points is used as the template and the other one as the template,
our method gives desired result (Fig. \ref{f5}) whereas the competing methods
give less accurate results (Supplementary Fig. S3).

\begin{figure}[!t]
  \centering
    \includegraphics[width=0.35\linewidth]{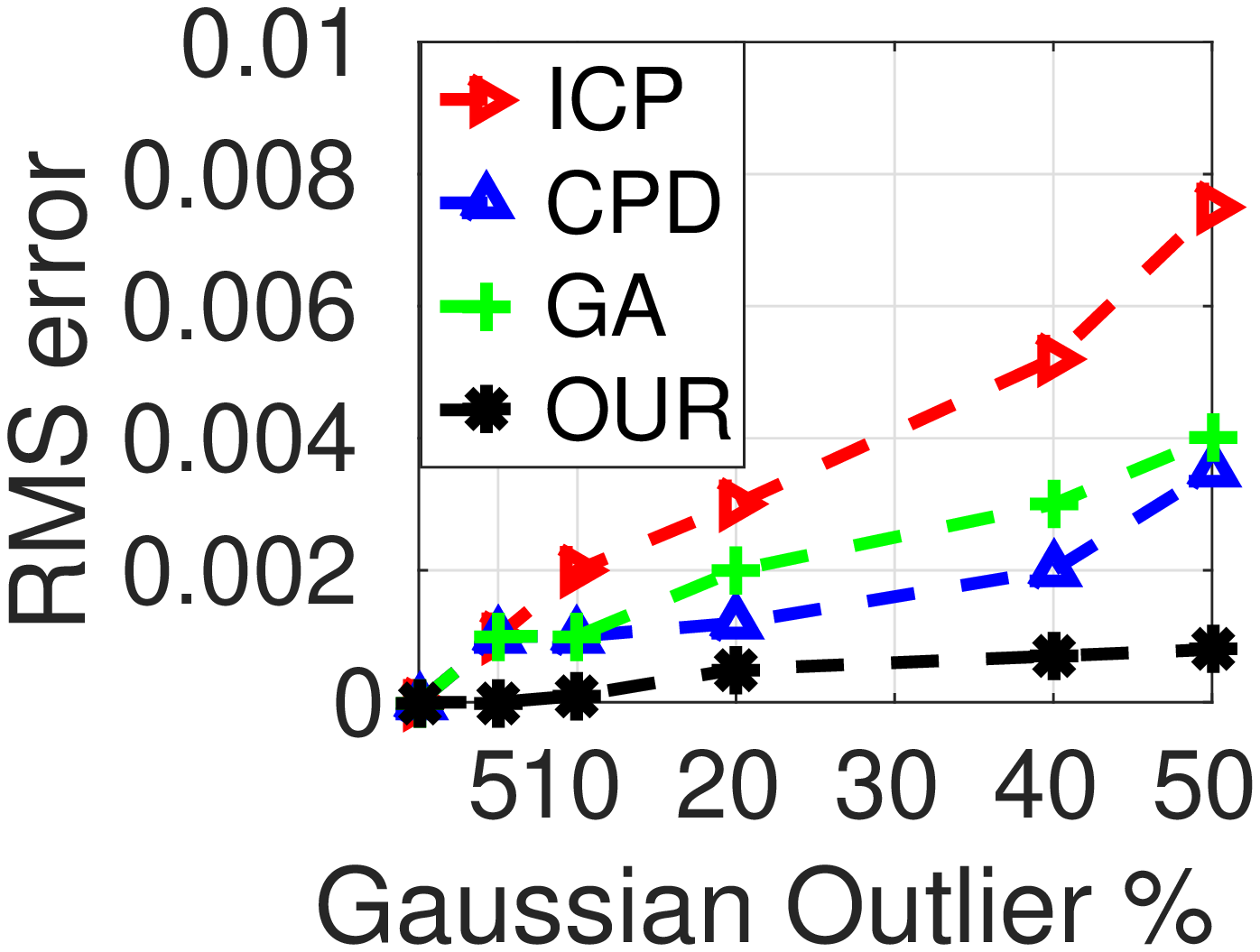}
    \includegraphics[width=0.35\linewidth]{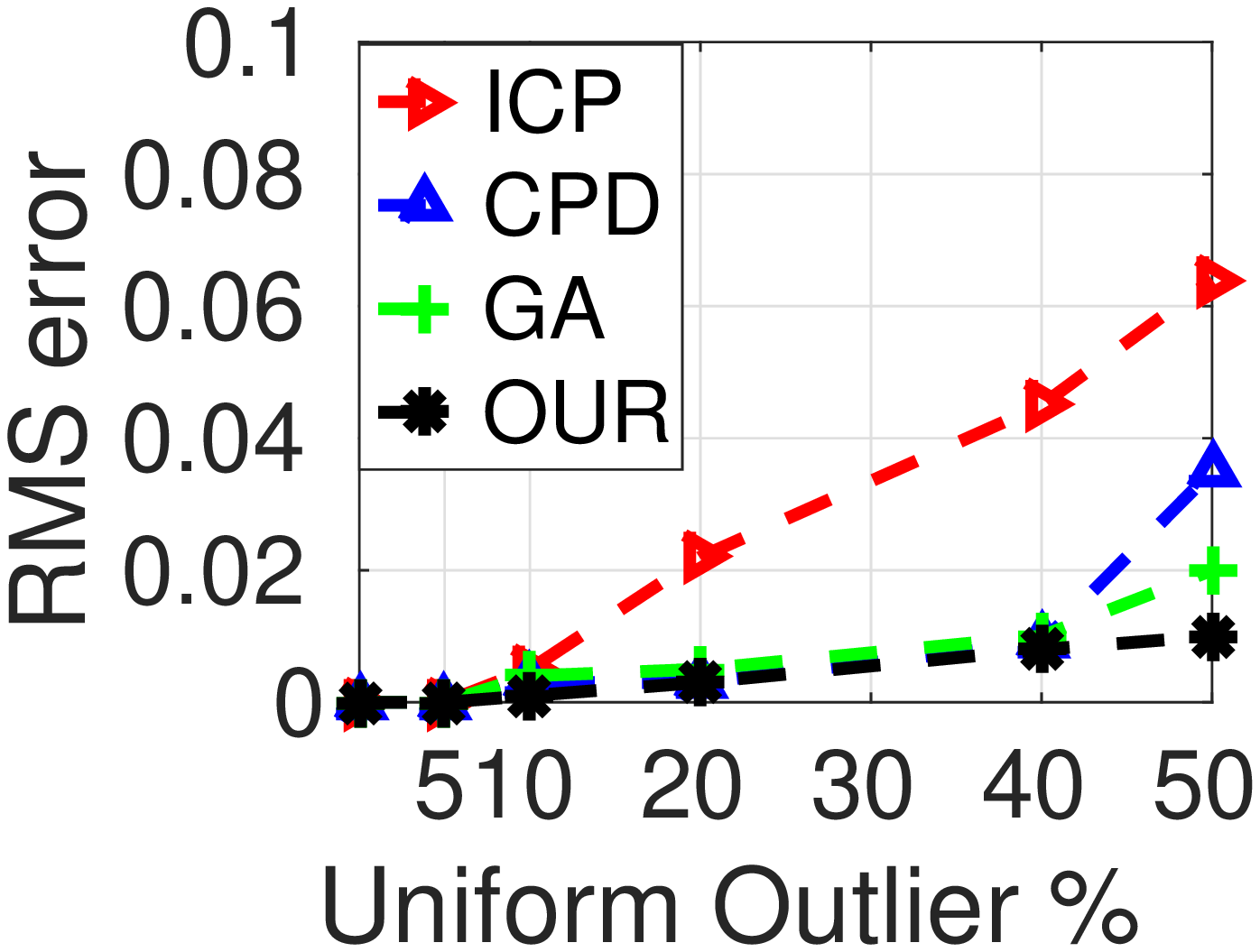}
 \vspace{-0.3cm}
\caption{Comparison of the proposed method with GA \cite{golyanik2016gravitational}, ICP \cite{besl1992method} and
CPD \cite{myronenko2010point} in the presence of unstructured outliers.}
\label{f7}
\end{figure}

Fig. \ref{f5}c and d show the results when 50\% Gausian and Uniform outliers respectively are added to the data. To 
quantitatively compare the proposed approach with that of 
GA \cite{golyanik2016gravitational}, CPD 
\cite{myronenko2010point} and ICP \cite{besl1992method}, we follow the same evalutaion protocol as used in 
\cite{golyanik2016gravitational} and use the same bunny point cloud from Stanford 3D Scanning Repository \cite{stanfordbunny}
as used in \cite{golyanik2016gravitational} and \cite{myronenko2010point} . We add 5\%, 10\%, 20\%, 40\% and 50\% Gaussian 
and uniform noise to
the template. For each of the 10 noise type-percentage combinations, 500 random tranformations are applied on the point cloud to
form the template. The non-transformed point-cloud acts as the reference. Fig. \ref{f7} compares the the amout of noise vs.
Root Mean Square Error (RMSE) for the four competing methods. All the methods perform better for Gaussian compared to uniform
noise. Overall CPD \cite{myronenko2010point} performs better compared to ICP \cite{besl1992method}. Our method outperforms all
the three competing methods.

\begin{figure}[!t]
  \centering
    \includegraphics[width=1.0\linewidth]{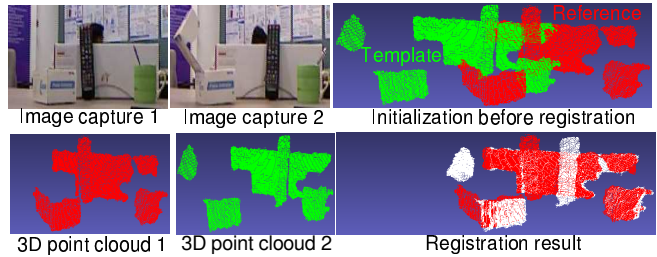}
 \vspace{-0.8cm}
\caption{Registration result on real scene.}
\label{f6}
\end{figure}

\textbf{On Real Data:} To evaluate the proposed method in real scenarios, we capture 3D point-cloud scans of a number of
scenes, two captures for each scene using Kinect. Fig. \ref{f6} demonstrates one example. In the second capture, 
the position of the remote-controller has been changed and the front box has been opened. To make the problem even more 
difficult, we translate
the template by 0.2 meters along x-axis. The registration result shows propoer alighment with the front box, cup and the background box.
Notice the cover of the front box and the remote-controller have not been aligned with anything as the postion of these 
items have been changed in the second capture. Thus, the proposed registration approach can be used for change detection in 
real scenes.
 \vspace{-0.2cm}
\section{Conclusions}
\label{VI}
 \vspace{-0.2cm}
We have proposed a novel rigid point-set registration algorithm with special characteristics (e.g., shape constraint, 
translation proportional to distance, spatial 
point-set distribution model for handling scale) that outperforms other competing approaches \cite{golyanik2016gravitational}, 
\cite{myronenko2010point} and \cite{besl1992method}. The proposed approach registers better in difficult conditions
such as missing object part, rotation more than $50^o$ and different amounts of structured and unstructured outliers. We
plan to employ the proposed registration approach for efficient change detection.

\bibliographystyle{IEEEbib}
\bibliography{Template}

\end{document}